# FPGA-based Controller for a Mobile Robot

Ms. Shilpa Kale

Dept. of Electronics & Telecommunication Engg.

Nagpur - 440025, India

E-mail : shilkale@rediffmail.com

Mr. S. S. Shriramwar

Dept. of Electronics & Telecommunication Engg.

Nagpur - 440025, India

E-mail: sshriramwar@yahoo.com

*Abstract*

**With application in the robotics and automation, more and more it becomes necessary the development of applications based on methodologies that facilitate future modifications, updates and enhancements in the original projected system. This project presents a conception of mobile robots using rapid prototyping, distributing the several control actions in growing levels of complexity and computing proposal oriented to embed systems implementation. This kind of controller can be tested on different platform representing the mobile robots using reprogrammable logic components (FPGA).**

**This mobile robot will detect obstacle and also be able to control the speed. Different modules will be Actuators, Sensors, wireless transmission. All this modules will be interfaced using FPGA controller. I would like to construct a mechanically simple robot model, which can measure the distance from obstacle with the aid of sensor and accordingly should able to control the speed of motor.**

**I would like to construct a mechanically simple robot model, which can measure the distance from obstacle with the aid of sensor and accordingly should able to control the speed of motor.**

## I. INTRODUCTION

The emergence of reconfigurable Field Programmable Gate Arrays (FPGA) has given rise to a new platform of complete mobile robot control system. With FPGA devices, it is possible to tailor the design to fit the requirements of applications (for example, exploration and navigation functions for a robot). General-purpose computers can provide acceptable performance when tasks are not too complex. A single processor system cannot guarantee real-time response (particularly in the absence of considerable additional hardware), if the environment is dynamic or semi-dynamic. This paper only focuses on the study of the mobile robot platform, with two driving wheels mounted on the same axis and a free front wheel. An FPGA-based robotic system can be designed to handle tasks in parallel. An FPGA-based robot also improves upon the single general purpose processor/computer based robot in the following areas:

1. Enhanced I/O channels. One can directly map the logical design to the computing elements in FPGA devices.

2. Low power consumption compared to desktops/laptops.

3. Support for the logical design of the non-Von Neumann computational models.

4. Support for easy verification of the correctness of the logical design modules.

Wheeled mobile robots (WMRs) are more energy efficient than legged or treaded robots on hard, smooth surfaces [Bekker60, Bekker691]; and will potentially be the first mobile robots to find widespread application in industry, because of the hard, smooth plant floors in existing industrial environments. WMRs require fewer and simpler parts and are thus easier to build than legged or treaded mobile robots. Wheel control is less complex than the actuation of multi-joint legs, and wheels cause minimal surface damage in comparison with treads.

The mobile robot consists of many units:

- mechanics (chassis, housing, wheels)
- electromechanical parts
- sensors

Robots carry out many various tasks. During these tasks the robot moves and orients. While navigating, it uses signals from the environment and the contents of its own memory to make the correct decisions. This form of navigation may be manifold depending on the given task and problem. Often the goal can be sensed, there is no obstacle between the goal and the robot, but there are numerous times when this is not the case, then the marking points must be sensed and the route known. In order for the robot to be able to do this, it must contain two main components:
• Drive, motion and
• control, direction.

## II. WHEELED MOBILE ROBOT

A robot capable of locomotion on a surface solely through the actuation of wheel assemblies mounted on the robot and in contact with the surface. A wheel assembly is a device which provides or allows relative motion between its mount and a surface on which it is intended to have a single point of rolling contact.

The simplest cases of mobile robots are wheeled robots, as shown in Figure 1. Wheeled robots comprise one or more driven wheels (drawn solid in the figure) and have optional passive or caster wheels (drawn hollow) and possibly steered wheels (drawn inside a circle). Most designs require two motors for driving (and steering) a mobile robot. The design is also steered. It requires two motors, one for driving the wheel on the left-hand side of Figure 1 has a single driven wheel that and one for turning. The advantage of this design is that the driving and turning actions have been completely separated by using two different motors. The robot design in the middle of Figure 1 is called "differential drive" and is one of the most commonly used mobile robot designs. The combination of two driven wheels allows the robot to be driven straight, in a curve, or to turn on the spot. Finally, on the right-hand side of Figure 1 is the so-called "Ackermann Steering", which is the standard drive and steering system of a rear-driven passenger car. We have one motor for driving both rear wheels via a differential box and one motor for combined steering of both front wheels.

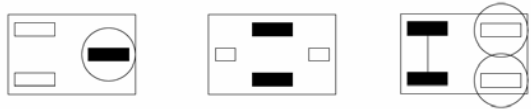

Figure 1: Wheeled Robots

## III. HARDWARE DESCRIPTION OF MOBILE ROBOT

### A. Architecture of Mobile Robot

Within the proposal of mobile robotics platform, the use of FPGA Controller, with control software especially developed for the necessary applications is considered using structured libraries to design, simulation, and verification with SIMULINK, we convert the model to function prototyping using FPGA hardware.

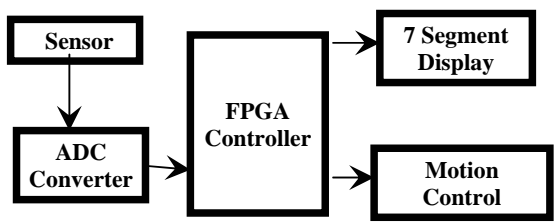

Figure 2: Block Diagram of mobile robot

The Figure 2 shows the overall flow of designing the system.

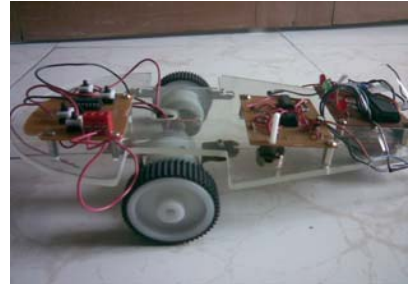

Figure 3: Mobile Robot Platform and elements

This system included both hardware and software development. The output of the ADC was connected to the FPGA board and it was used as the input of the source code. The assembly language that used in this system is Verilog HDL. After the simulation and the synthesis process, the program has been implemented on the FPGA board.

Figure 3. shows mobile robot platform and its elements.

### B. Interfacing the FPGA with the L293D for Motor Control

The FPGA will process the PWM program and the output will be given to enable pin of L293D which activate the L293 quadruple high current half H-Driver chip (L293 Datasheet) and controls the speed of the motor. Table 1 shows the truth table to get the L293 to perform the different movement operations such as "Forward" etc.

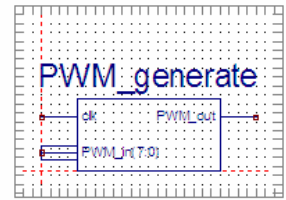

Figure 4: PWM schematic symbol

| FUNCTIONS | INPUTS (ENA ENB 1A 2A 3A 4A) |
|-----------|------------------------------|
| Forward | 111010 |
| Reverse | 110101 |
| Left | 010010 |
| Right | 101000 |

**Table 1: Logic inputs to activate the L293 chip**

### C. Interfacing the FPGA with ADC0809

An A/D converter translates an analog signal into a digital value. An 8 channel, 8-bit A/D input is available to read analog voltages between 0 to 5 Volts. Devices such as an analog joystick or potentiometers can be connected to one of ADC channel and converted digital output can be read and is sent back to the FPGA board to control the speed of DC motor. Figure 5 shows the construction of the ADC. The characteristics of an A/D converter include:
• Accuracy
expressed in the number of digits it produces per value (for example 10bit A/D converter)
• Speed
expressed in maximum conversions per second
(for example 500 conversions per second)
• Measurement range
expressed in volts (for example 0.5V)

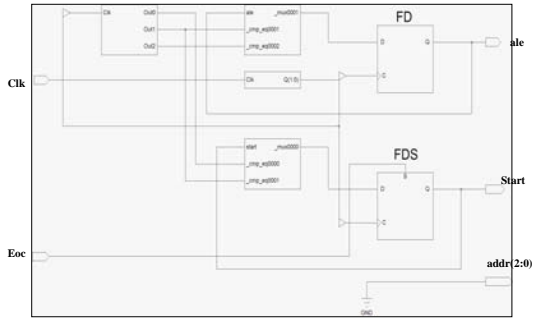

Figure 5: ADC system shown in the Xilinx-ISE environment

*1) Distance Measurement Sensor*

The analog sensor Sharp GP2D12 simply returns a voltage level in relation to the measured distance. In Figure 6, above, the relationship between digital sensor read-out (raw data) and actual distance information can be seen. From this diagram it is clear that the sensor does not return a value linear or proportional to the actual distance, so some post-processing of the raw sensor value is necessary. The simplest way of solving this problem is to use a lookup table which can be calibrated for each individual sensor.

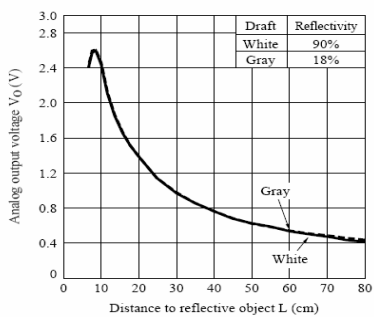

Figure 6: Analog output voltage Vs Distance to Reflective Objects

### IV. RESULTS

*A. Synthesis Results*

The full VHDL system consists of all the previous blocks connected together, and simulated as a whole before they were downloaded to the FPGA. Figure 8 shows the simulation results obtained from the ADC block. Here START and ALE signals must be high for atleast 100ns to start the conversion process of ADC converter. When conversion is completed, then EOC & OE signals will be pulsed high. The output in digital form is then given to FPGA as an input. Depending on digital input, PWM output will be calculated. Figure 9 shows simulation result of PWM block.

Other simulations results are of seven segment display as shown in figure 10 which display the correct numeral numbers, when adc_in[0:3] input given is in between 0-9 i.e; in binary "0000" to "1001". It displays the measured distance in cm by GP2D12 sensor. Figure 11 shows Prototype board built with FPGA XC2S50 with simulation result as shown in figure 12.

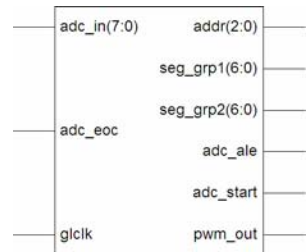

Figure 7: RTL Schematic of Mobile Robot

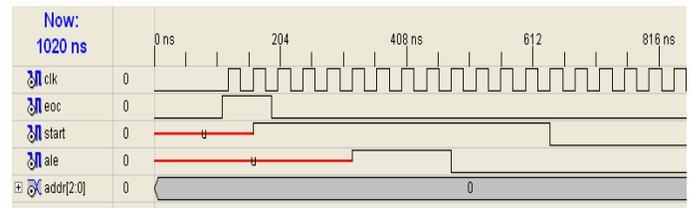

Figure 8: Simulation result of ADC VHDL code

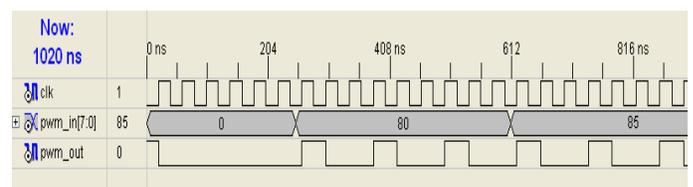

Figure 9: Simulation result of PWM VHDL code

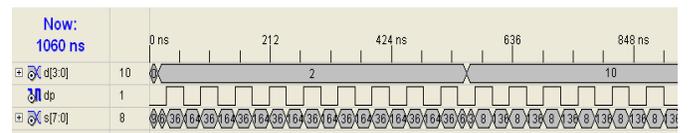

Figure 10: Simulation result of 7-segment VHDL code

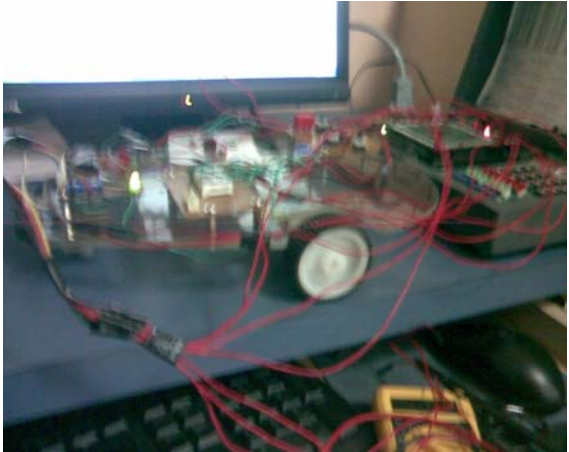

Figure 11: Prototype board built with FPGA XC2S50

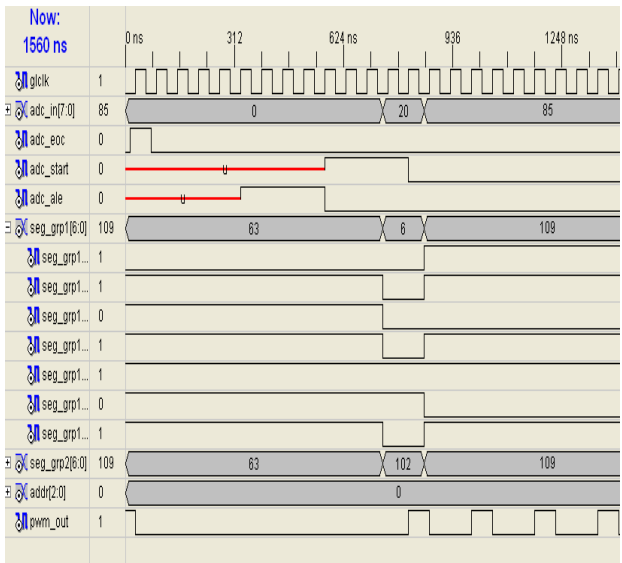

Figure 12 : Simulation result of Full system VHDL code

## V. Summary

This paper describes the design and implementation of a VHDL code for a simple mobile robot. This design may be described as a mapping from the input sensors to the actuators which control the robot motions. It is shown that FPGA can be configured to implement the design successfully.

This paper had been discussed about the methods that used in this project. There are four phases in completing this project. The phase one is design entry which is including both software and hardware development. The hardware developments are the driver circuit for motor and the ADC circuit and the software that being used is ISE Simulator. The simulation process is used to verify the design and the synthesis process is used to produce the block diagram.

## VI. Conclusions & Future work

Robotic platforms engineering are a necessary in teaching and research institutions for knowledge consolidation in several teaching and research, such as modeling, control, automation, power systems, embedded electronics and software. The use of the mobile robots for this purpose appears to be quite an attractive solution. It allows the integration of several important areas of knowledge and a low cost solution, which has already been adopted with success by other research institutions. The main objective of this work was to propose a generic platform for a mobile robot system, seeking to obtain a support tool for under-graduation and graduation activities. Another objective was to gather knowledge in the mobile robotic area, aiming at presenting practical solutions for industrial problems, such as maintenance, supervision and transport of materials.

The mobile robot systems have more and more importance these days, so dealing with them in the higher education is necessarily. Autonomous mobile robots can be used to deliver parts in factories, being complementary platforms in a security system and they also can be used in hazardous areas where humans can not stay.

The wireless channel may also be added to increase system flexibility. The proposed framework remains simple and user friendly; additionally it provides enough flexibility for the specific application. Our approach can be extended to more demanding applications by adding more modules, or other peripheral interfaces. Currently I am working on the development of VHDL code which integrates all modules which are discussed above.

## Acknowledgement

I would like to express my acknowledgement for the support and resources for this project that have been obtained from Prof. S. S. Shriramwar, all from the Priyadarshani College of Engg, Nagpur University. I would like to thank my Guide for his help in setting up the design system and for his expert advice on FPGA.